%% file: paper.tex
\documentclass[runningheads]{llncs}

\usepackage{silence}
\usepackage{includes/eccv}

\usepackage{includes/eccvabbrv}

\usepackage{graphicx}
\usepackage{booktabs}

\usepackage[accsupp]{axessibility}  

\usepackage{xr-hyper}
\makeatletter

\newcommand*{\addFileDependency}[1]{
\typeout{(#1)}
\@addtofilelist{#1}
\IfFileExists{#1}{}{\typeout{No file #1.}}
}\makeatother

\usepackage{hyperref}

\usepackage{orcidlink}

\AddToHook{cmd/appendix/before}{%
  \setcounter{equation}{0}%
  \setcounter{theorem}{0}%
}
\usepackage{mathtools}
\usepackage{dashbox}
\usepackage{mathtools, amssymb, amsmath}
\usepackage{algorithm}
\usepackage[noend]{algorithmic}
\usepackage{booktabs}
\usepackage{multirow}
\usepackage{wrapfig}
\usepackage{tikz}

\usetikzlibrary{arrows.meta}
\definecolor{myblue}{HTML}{8aa9c6}
\newcommand{\bluearrow}[1]{\begin{tikzpicture}\draw[-{Latex[width = 2mm, length=1mm]}, line width=#1, color = myblue] (0,0.2) -- (0,0);\end{tikzpicture}}
\newcommand{\smallbluearrow}[1]{\begin{tikzpicture}\draw[-{Latex[width = 1mm, length=1mm]}, line width=#1, color = myblue] (0,0.2) -- (0,0);\end{tikzpicture}}
\newcommand{\bigbluearrow}[1]{\begin{tikzpicture}\draw[-{Latex[width = 3mm, length=1mm]}, line width=#1, color = myblue] (0,0.2) -- (0,0);\end{tikzpicture}}
\newcommand{\grayarrow}[1]{\begin{tikzpicture}\draw[-{Latex[width = 2mm, length=1mm]}, line width=#1, color = gray] (0,0) -- (0.2,0);\end{tikzpicture}}

\newcommand{\method}{CITRUS\xspace}

\DeclareMathOperator*{\argmax}{arg\,max}
\DeclareMathOperator*{\argmin}{arg\,min}
\newcommand{\din}{d_{\text{in}}}
\newcommand{\dout}{d_{\text{out}}}

\definecolor{mygreen}{HTML}{B4E5A2}
\definecolor{myred}{HTML}{EEA3A3}
\definecolor{mydarkred}{HTML}{C00000}
\newcommand\dashedbox[1][H]{\setlength{\fboxsep}{0pt}\setlength{\dashlength}{2.2pt}\setlength{\dashdash}{1.1pt} \dbox{\phantom{#1}}}
\newcommand\greenbox[1][H]{\setlength{\fboxsep}{0.3pt}\setlength{\dashlength}{2.2pt}\setlength{\dashdash}{1.1pt} \colorbox{mygreen}{\phantom{#1}}}
\newcommand\redbox[1][H]{\setlength{\fboxsep}{0.3pt}\setlength{\dashlength}{2.2pt}\setlength{\dashdash}{1.1pt} \colorbox{myred}{\phantom{#1}}}
\newcommand\darkredbox[1][H]{\setlength{\fboxsep}{0.3pt}\setlength{\dashlength}{2.2pt}\setlength{\dashdash}{1.1pt} \colorbox{mydarkred}{\phantom{#1}}}

\newenvironment{sproof}{%
  \proof}{\endproof}

\begin{document}

\title{Cross-Input Certified Training for Universal Perturbations} 

\titlerunning{Certified Training for UPs}

\author{Changming Xu\inst{1}\orcidlink{0000-0003-3079-5652} \and
Gagandeep Singh\inst{1,2}\orcidlink{0000-0002-9299-2961}}

\authorrunning{C.~Xu et al.}

\institute{University of Illinois Urbana Champaign, Champaign IL 61820, USA \and
VMWare Research, Palo Alto CA 94304, USA}

\maketitle

\begin{abstract}
Existing work in trustworthy machine learning primarily focuses on single-input adversarial perturbations. In many real-world attack scenarios, input-agnostic adversarial attacks, e.g. universal adversarial perturbations (UAPs), are much more feasible. Current certified training methods train models robust to single-input perturbations but achieve suboptimal clean and UAP accuracy, thereby limiting their applicability in practical applications. We propose a novel method, \method, for certified training of networks robust against UAP attackers. We show in an extensive evaluation across different datasets, architectures, and perturbation magnitudes that our method outperforms traditional certified training methods on standard accuracy (up to 10.3\%) and achieves SOTA performance on the more practical certified UAP accuracy metric.
\end{abstract}

\input{contents/1-introduction}

\input{contents/2-background}
\input{contents/3-technical_contribution}

\input{contents/4-evaluation}
\input{contents/5-related_work}
\input{contents/6-conclusion}

%
%
\bibliographystyle{includes/splncs04}
\bibliography{main}

\newpage
\input{contents/7-appendix}
\end{document}

%% file: contents/1-introduction.tex
\section{Introduction}

Over the last few years, deep neural networks (DNNs) have found widespread application in safety-critical systems (autonomous driving \cite{shafaei2018uncertainty, grigorescu2020survey}, medical diagnosis \cite{kononenko2001machine, finlayson2019adversarial}, etc.), emphasizing the need for their robustness and reliability. The reliability of DNNs has been especially challenged by adversarial examples \cite{fgsm, pgd, cw, turtle, uap}, inputs which have been maliciously modified to cause misclassification. To counteract this, many recent works have proposed methods for formally verifying safety of DNNs \cite{deeppoly, marabou, neurify, planet} and for training DNNs to have stronger formal guarantees \cite{wong, gowal2018effectiveness, crownibp, colt, sabr, taps, mirman2018differentiable, yang2023provable}. However, despite the many advancements in this area, current certified training methods achieve suboptimal clean accuracy and focus primarily on single-input adversarial perturbations.

A majority of adversarial perturbations work is focused on computing an $l_p$ bounded single-input additive perturbation \cite{pgd, fgsm, cw}. In many practical scenarios, attackers can not feasibly compute and apply single-input adversarial perturbations in real-time. Recent work has shown input-agnostic attacks, specifically, universal adversarial perturbations (UAPs) to be a more realistic threat model \cite{music:19, sticker:19, wruap}. UAPs have no dependence on single inputs, instead they are learned to affect \textit{most} inputs in a data distribution. For these scenarios, it is overly conservative to assume the single-input adversarial region model for verification/certified training. Existing work \cite{uat, cwuat, perolat2018playing, wruap} on training networks to be empirically robust to UAPs suggests that focusing on this more realistic threat model leads to improved accuracy.

\noindent\textbf{Challenges}. The training problem for the UAP robustness (\cref{sec:uaptr}) involves maximizing the expected loss under a single perturbation applied to multiple inputs. The maximization requires relational cross-executional reasoning; applying the same perturbation to multiple inputs. This is harder to optimize than the standard adversarial robustness problem (\cref{sec:tr}) which is the expectation of maximum loss. The maximization in standard robustness can be broken down into simpler single-input optimization problems. 

\noindent \textbf{This work.} We propose the \textit{first certified training method} for universal perturbations, \textbf{C}ross-\textbf{I}nput certified \textbf{TR}aining for \textbf{U}niversal perturbation\textbf{S} (\method), for robustness to UAP-based adversaries. Our method is based on the following \textit{key insight}: training on cross-input adversarial sets satisfies the UAP robustness objective and leads to less regularization and higher standard accuracy than training on single-input adversarial sets.

\noindent\textbf{Main Contributions}. Our main contributions are:

\begin{itemize} 
    \item A UAP robustness training objective and theoretical investigation on a set of upper bounds for the UAP loss based on maximizing over sets of common perturbations (\cref{sec:certuap}).
    \item We develop \method, the first certified training method for UAP robustness designed using our upper bounds (\cref{sec:method}).
    \item A thorough evaluation showing that \method obtains better accuracy (up to 10.3\% higher) and SOTA certified UAP accuracy compared to existing certified training methods \cite{sabr, taps,shiibp} on popular datasets for certified training such as CIFAR10 and TinyImageNet. We further show that against SOTA universal adversarial training methods \cite{uat,cwuat}, \method obtains much more certified UAP accuracy (35.62\% compared to 1.51\%). 
\end{itemize}

%% file: contents/2-background.tex
\section{Background} \label{sec:background}

In this section, we provide the necessary background for \method.

\subsection{Adversarial Perturbations and Robustness}
Given an input-output pair $(\mathbf{x},y) \in \mathcal{X} \subseteq \mathbb{R}^{\din}\times\mathbb{Z}$, and a classifier $f: \mathbb{R}^{\din} \to \mathbb{R}^{\dout}$ which gives score $f_k(\mathbf{x})$ for class $k$ (let $\hat{f}(\mathbf{x}) = \argmax_k f_k(x)$). An additive perturbation, $\mathbf{v} \in \mathbb{R}^{\din}$, is adversarial for $f$ on $\mathbf{x}$ if $\hat{f}(\mathbf{x} + \mathbf{v}) \neq y$. A classifier is adversarially robust on $\mathbf{x}$ for an $l_p$-norm ball, $\mathcal{B}_p(0, \epsilon)$, if it classifies all elements within the ball added to $\mathbf{x}$ to the correct class. Formally, $\forall \mathbf{v} \in \mathcal{B}_p(0, \epsilon). \hat{f}(\mathbf{x} + \mathbf{v}) = y$. In this paper, we focus on $l_\infty$-robustness, i.e. balls of the form $\mathcal{B}_\infty(\mathbf{x}, \epsilon) \coloneqq \{\mathbf{x'} = \mathbf{x} + \mathbf{v} | \|\mathbf{v}\|_\infty \leq \epsilon\}$, so will drop the subscript $\infty$. Note, our method can be extended to other $l_p$ balls.

\subsection{Universal Adversarial Perturbations}
An additive perturbation $\mathbf{u} \in \mathbb{R}^{\din}$ is considered a UAP if it has a high probability of being adversarial for all elements in an input distribution. Formally, let $p(\mathbf{u}) \coloneqq P_{\mathbf{x} \in \mathcal{X}}[\hat{f}(\mathbf{x} + \mathbf{u}) \neq \hat{f}(\mathbf{x})]$, then $\mathbf{u}$ is a UAP with threshold $p(\mathbf{u})$. We can define worst-case UAP robustness over a data distribution, $\mathcal{X}$, and ball $\mathcal{B}(0, \epsilon)$ as the highest UAP threshold, $\gamma^\ast$, for any vector in the ball. Formally, we have $\gamma^\ast = \min(\gamma) \text{ s.t. } \forall \mathbf{u} \in \mathcal{B}(0, \epsilon). p(\mathbf{u}) \leq \gamma$.

\subsection{Neural Network Verification}
Above we introduced the idea of adversarial robustness, verification is the process of formally proving robustness properties. The proportion of inputs that we can prove robustness on is denoted as the certified accuracy of a network.  One popular and effective method is interval bound propagation (IBP) or \textsc{Box} propagation \cite{mirman2018differentiable, gowal2018effectiveness}. IBP first over-approximates the input region, $b^0 = \mathcal{B}(\mathbf{x}, \epsilon)$, as a \textsc{Box} $[\underline{\mathbf{x}}^0, \overline{\mathbf{x}}^0]$ where each dimension is an interval with center $\mathbf{x}$ and radius $\epsilon$. Let our network, $f = L_1 \circ L_2 \circ \dots \circ L_n$, be the composition of $n$ linear or ReLU activation layers (for this paper we consider networks of this form, although our methods could be extended to other architectures). We then propagate the input region $b^0$ through each layer. For details of this propagation see \cite{mirman2018differentiable, gowal2018effectiveness}. At the output layer, we would like to show that the lower bound of the true class is greater than all upper bound of all other classes, i.e. $\forall i \in \dout, i\neq y. \overline{\mathbf{o}}_i - \underline{\mathbf{o}}_y < 0$. Further verification methods are mentioned in \cref{sec:relatedwork}.

\subsection{Training for Robustness}\label{sec:tr}
For single-input robustness, we minimize the expected worst-case loss due to adversarial examples leading to the following optimization problem \cite{sabr, taps, pgd}: 

\begin{equation}\label{eq:trainrob}
    \mathbf{\theta} = \argmin_{\mathbf{\theta}} \mathop{\mathbb{E}}_{(\mathbf{x}, y) \in \mathcal{X}}\left[ \max_{\mathbf{x'} \in \mathcal{B}(\mathbf{x}, \epsilon)} \mathcal{L}(f_\theta(\mathbf{x}'), y)\right]
\end{equation}

Where $\mathcal{L}$ is a loss over the output of the DNN. Exactly solving the inner maximization is computationally impractical, in practice, it is approximated. Underapproximating the inner maximization is typically called adversarial training, a popular technique for obtaining good empirical robustness \cite{pgd}, but these techniques do not give formal guarantees and are potentially vulnerable to stronger attacks \cite{tramer2020adaptive}. We will focus on the second type of training, called certified training which overapproximates the inner maximization.

\subsection{Certified Training}
The IBP verification framework above adapts well to training. The \textsc{Box} bounds on the output can be encoded nicely into a loss function:

\begin{equation}
    \mathcal{L}_{\text{IBP}} (\mathbf{x}, y, \epsilon) \coloneqq \ln\left(1 + \sum_{i\neq y} e^{\overline{\mathbf{o}}_i - \underline{\mathbf{o}}_y} \right)
\end{equation}

To address the large approximation errors arising from \textsc{Box} analysis, SABR \cite{sabr}, a SOTA certified training method, obtains better standard and certified accuracy by propagating smaller boxes through the network. They do this by first computing an adversarial example, $\mathbf{x}' \in \mathcal{B}(\mathbf{x}, \epsilon - \tau)$ in a slightly truncated $l_\infty$-norm ball. They then compute the IBP loss on a small ball around the adversarial example, $\mathcal{B}(\mathbf{x}', \tau)$, rather than on the entire ball, $\mathcal{B}(\mathbf{x}, \epsilon)$, where $\tau \ll \epsilon$. 

\begin{equation}
    \mathcal{L}_{\text{SABR}} (\mathbf{x}, y, \epsilon, \tau) \coloneqq \max_{x' \in \mathcal{B}(\mathbf{x}, \epsilon - \tau)} \mathcal{L}_{\text{IBP}} (\mathbf{x}', y, \tau)
\end{equation}

Even though this is not a sound approximation of adversarial robustness, SABR accumulates fewer approximation errors due to its more precise \textsc{Box} analysis; thus, reduces overregularization and improves standard/certified accuracy.

%% file: contents/3-technical_contribution.tex
\section{Certified Training for Universal Perturbations} \label{sec:certuap}

In this section, we introduce our formal training objective for certified UAP robustness. This objective is hard to compute so we instead seek to bound this objective. Lower bounds are akin to adversarial training and, as we show, in \cref{sec:ablation} do not obtain good certified accuracy against universal perturbations. Therefore, we define upper bounds that can be efficiently computed. These form the backbone for the \method algorithm.

\subsection{UAP Training Objective}\label{sec:uaptr}

\Cref{eq:trainrob} gives the training objective for standard single-input adversarial robustness: minimizing the expected loss over the data distribution due to worst-case adversarial perturbations crafted separately for each input. For UAP robustness, we minimize the worst-case expected loss from a single perturbation applied to all points in the data distribution.

\begin{equation} \label{eq:uaptrainrob}
    \theta = \argmin_\theta \max_{\mathbf{u}\in \mathcal{B}(0,\epsilon)}\left(\mathop{\mathbb{E}}_{(x,y) \in \mathcal{X}}\left[ \mathcal{L}(f_\theta(\mathbf{x} + \mathbf{u}), y)\right]\right)
\end{equation}

For this paper we assume that $\mathcal{L}$ refers to a standard loss function where adversarial additive perturbations incur greater loss than safe perturbations. Since UAPs are input-agnostic we maximize the expected value over $\mathbf{u}\in\mathcal{B}(\mathbf{0},\epsilon)$. Solving this maximization exactly is computationally impractical \cite{uap, perolat2018playing}. To create an efficient training algorithm for certified UAP robustness we need an efficiently computable upper bound for the maximization. It is necessary to reduce the approximation error due to the upper bound as existing research has shown that training with a looser upper bound leads to more regularization and a reduction in clean accuracy \cite{sabr}. 

\subsection{$k$-Common Perturbations}

In this section, we introduce the idea of $k$-common perturbations ($k$-cp) and show that the maximization of \cref{eq:uaptrainrob} can be upper bounded by the expected loss over the dataset due to the worst-case $k$-cp (\Cref{th:iter}). By doing this, we get a set of losses based on $k$-cps. In \cref{sec:method} we use the upper bounds derived in this section to create an algorithm for certified UAP training which we show experimentally (\cref{sec:eval}) reduces regularization and increases accuracy while also achieving certified UAP accuracy.

We first define a boolean predicate, $A_f:\mathbb{R}^{\din}\times\mathbb{N} \to \{\textit{true}, \textit{false}\}$ which is true when $f$ is adversarial for $\mathbf{x}, y$:

\begin{equation*}
A_f(\mathbf{x}, y) = 
\begin{cases}
    \textit{true} &\text{ if } \hat{f}(\mathbf{x}) \neq y\\
    \textit{false} &\text{ otherwise}
\end{cases}
\end{equation*}

\begin{definition}
 $\mathbf{u}\in\mathbb{R}^{\din}$ is a $k$-common perturbation on a data distribution $\mathcal{X}$ for a network $f$, if there exists a set of $k$ inputs for which $\mathbf{u}$ is adversarial. That is, if $\exists \{(\mathbf{x}_i, y_i) |i\in [k]\} \subseteq \mathcal{X}. \bigwedge_{i=1}^k A_f(\mathbf{x}_i + \mathbf{u}, y_i)$.   
\end{definition}
Note that if $\mathbf{u}$ is a $k$-cp then $\mathbf{u}$ is also a $(k-1)$-cp, $\dots$, $1$-cp. Next, we define a boolean predicate function $\Psi_{\mathcal{X}, f}: \mathbb{R}^{\din}\times\mathbb{N}\to \{\textit{true}, \textit{false}\}$ which computes whether a given perturbation $\mathbf{u}$ is a $k$-cp w.r.t. $\mathcal{X}$ and $f$

\begin{equation*}
\Psi_{\mathcal{X}, f}(\mathbf{u},k) = 
\begin{cases}
    \textit{true} & \text{ if } \exists \{(\mathbf{x}_i,y_i)|i\in [k]\} \subseteq \mathcal{X}.\bigwedge_{i=1}^k A_f(\mathbf{x}_i + \mathbf{u}, y_i)\\
    \textit{false} & \text{ otherwise}
\end{cases}
\end{equation*}

Since $\Psi$ is monotonic w.r.t. $k$, there is a unique $j$ for each $\mathbf{u}$ where $\Psi_{\mathcal{X}, f}(\mathbf{u}, j)$ transitions from $1\to 0$, i.e. $\exists j\in\mathbb{N}$ s.t. $\forall i\leq j. \Psi_{\mathcal{X}, f}(\mathbf{u}, i)$ and $\forall i>j. \Psi_{\mathcal{X}, f}(\mathbf{u}, i) $.  

\begin{definition}
$\hat{\Psi}_{\mathcal{X}, f}: \mathbb{R}^{\din}\to\mathbb{N}$ is a function which computes the transition point of $\Psi_{\mathcal{X}, f}$ for $\mathbf{u}$. That is, $\hat{\Psi}_{\mathcal{X}, f}(\mathbf{u}) = \argmax_{k\in\mathbb{N}} k\cdot \Psi_{\mathcal{X}, f}(\mathbf{u}, k)$.    
\end{definition}

For ease of notation, we refer to $\Psi(\cdot) \coloneq \Psi_{\mathcal{X}, f}(\cdot)$ and $\hat{\Psi}(\cdot) \coloneq \hat{\Psi}_{\mathcal{X}, f}(\cdot)$.

\begin{definition}
   A $k$-cp set, $\mathcal{C}_{\mathcal{X}, f}(k, \epsilon)$, is the set of all perturbations in $\mathcal{B}(\mathbf{0}, \epsilon)$ that are $k$-cps w.r.t. $\mathcal{X}$ and f. That is, $\mathcal{C}_{\mathcal{X}, f}(k, \epsilon) = \{\mathbf{u} | \mathbf{u} \in \mathcal{B}(\mathbf{0}, \epsilon), \Psi(\mathbf{u}, k)\}$.
\end{definition}

Let $\mathbf{u}^\ast \in \mathcal{B}(\mathbf{0}, \epsilon)$ be the point at which the expectation of \cref{eq:uaptrainrob} is maximized (without loss of generality, we assume $\mathbf{u}^\ast$ is unique). Formally,

\begin{equation}\label{eq:uast}
\mathop{\mathbb{E}}_{(x,y) \in \mathcal{X}}\left[ \mathcal{L}(f(\mathbf{x} + \mathbf{u}^\ast), y)\right] = \max_{\mathbf{u}\in \mathcal{B}(0,\epsilon)}\left(\mathop{\mathbb{E}}_{(x,y) \in \mathcal{X}}\left[ \mathcal{L}(f(\mathbf{x} + \mathbf{u}), y)\right]\right)
\end{equation}

If $\mathbf{u}^\ast$ is a $k$-cp then the loss for an individual input when perturbed by $\mathbf{u}^\ast$ is bounded by maximizing the loss for that input over the entire $k$-cp perturbation set (Lemma 
1). We also know that all $j$-cps are $k$-cps for $j<k$ so $\mathcal{C}(j, \epsilon) \subseteq \mathcal{C}(k,\epsilon)$ (Lemma 
2). These two lemmas allow us to bound $\mathcal{L}(f(\mathbf{x} + \mathbf{u}^\ast), y)$ for each pair $(\mathbf{x}, y)$. Formal statements and proofs of these lemmas can be seen in Appendix 
A.1. We can now show that the maximization in \cref{eq:uaptrainrob} can be upper bounded by an increasing sequence of values. Each value is an expectation, over the full data distribution, of the worst-case loss achievable by a $k$-cp. Formally,

\begin{theorem}\label{th:iter}
    Given $\mathcal{X}\subseteq \mathbb{R}^{\din}\times\mathbb{N}$, network $f:\mathbb{R}^{\din}\to\mathbb{R}^{\dout}$, $\mathbf{u}^\ast$ as defined in \cref{eq:uast}, and norm-bound $\epsilon \in \mathbb{R}$. Let $\kappa^\ast = \hat{\Psi}_{\mathcal{X}, f}(\mathbf{u}^\ast)$ and $\mathcal{E}(k, \epsilon) = \mathop{\mathbb{E}}_{(x,y) \in \mathcal{X}}\left[ \max_{\mathbf{u}\in\mathcal{C}_{\mathcal{X}, f}(k, \epsilon)}\mathcal{L}(f(\mathbf{x} + \mathbf{u}), y)\right]$, then
    \begin{equation*}
        \max_{\mathbf{u}\in \mathcal{B}(\mathbf{0},\epsilon)}\left(\mathop{\mathbb{E}}_{(\mathbf{x},y) \in \mathcal{X}}\left[ \mathcal{L}(f(\mathbf{x} + \mathbf{u}), y)\right]\right)\leq \mathcal{E}(\kappa^\ast, \epsilon) \leq \mathcal{E}(\kappa^\ast - 1, \epsilon) \leq \dots \leq \mathcal{E}(1, \epsilon)
    \end{equation*}
\end{theorem}
\begin{sproof}
LHS equals $\mathop{\mathbb{E}}_{(x,y) \in \mathcal{X}}\left[ \mathcal{L}(f(\mathbf{x} + \mathbf{u}^\ast), y)\right]$ via Lemma 
5 which is upperbounded by $\mathcal{E}(\kappa^\ast, \epsilon)$ by Lemma 
1 applied to each element in the distribution. $\mathcal{E}(\kappa^\ast, \epsilon)$ is upperbounded by $\mathcal{E}(\kappa^\ast-1, \epsilon), \dots, \mathcal{E}(1, \epsilon)$ by Lemma 
2 applied to each element in the distribution. Full proof is in Appendix 
A.1.
\end{sproof}


\section{\method} \label{sec:method}

In this section, we will leverage \Cref{th:iter} to develop our \method algorithm for certified training against UAPs.

\subsection{$k$-CP Guided Losses} \label{sec:setloss}

\Cref{th:iter} gives us a sequence of upper bounds for the maximization of the UAP robustness objective. We will now focus on a batch-wise variant of these upper bounds for training. Given a batch of inputs $\mathcal{X}_B \subseteq \mathbb{R}^{\din}\times\mathbb{Z}$, current input $(\mathbf{x}, y) \in \mathcal{X}_B$, and for each $k \in [\kappa^\ast]$ we get the following loss functions

\begin{equation}
    \mathcal{L}_{\text{$k$CP}}(\mathcal{X}_B, \mathbf{x}, y, \epsilon) = \max_{\mathbf{u}\in\mathcal{C}_{\mathcal{X}, f}(k, \epsilon)}\mathcal{L}(f(\mathbf{x} + \mathbf{u}), y)
\end{equation}

However, in practice, computing $\kappa^\ast$ is intractable as it would either require computing individual adversarial regions \cite{dimitrov2021provably} and intersecting them or require finding $\mathbf{u}^\ast$ \cite{uap}. $\mathcal{L}_{\text{$k$CP}}$ is only an upper bound when $k \leq \kappa^\ast$ as if $k > \kappa^\ast$ then $\kappa^\ast$ does not fall in the $k$-cp region. 

Ideally, we would use the tightest upper bound ($\mathcal{L}_{\kappa^\ast\text{CP}}$), to induce the least amount of overapproximation; however, in practice, we do not know $\kappa^\ast$. For this paper, we consider $\mathcal{L}_{\text{$2$CP}}$ as it leads to efficient training algorithms. $\mathcal{L}_{\text{$1$CP}}$ is akin to standard certified training for single-input adversarial perturbations. Current research shows that networks trained by SOTA standard certified training can not eliminate single-input adversarial perturbations altogether \cite{sabr, taps, shiibp}, i.e. $\kappa^\ast \geq 1$ for these networks. Thus, when training networks to be certifiably robust to UAPs it is safe to assume that $\kappa^\ast \geq 1$. $\mathcal{L}_{\text{$2$CP}}$ penalizes perturbations affecting multiple inputs while still having a high chance of being an upper bound to the UAP robustness problem (if $\kappa^\ast = 1$ then we likely do not have any adversarial perturbations which affect multiple inputs and have succeded in training a network certifiably robust to UAPs). 

\subsection{\method Loss}\label{sec:citrusloss}

\begin{figure*}[t]
    \centering
\includegraphics[width=1\textwidth]{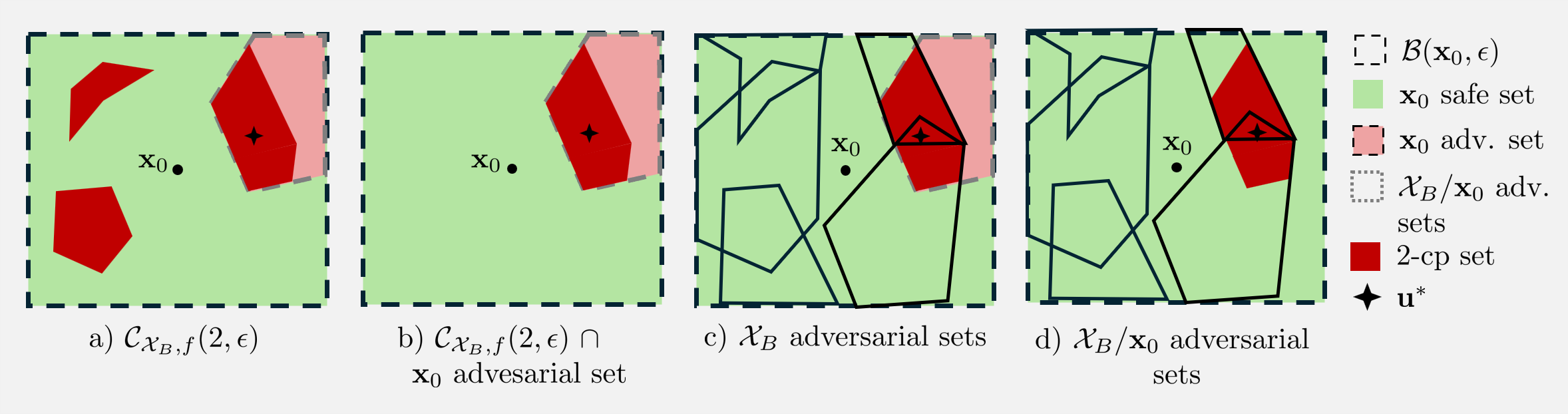}
    \caption{Upper bounding $\mathcal{L}_{2\text{CP}}$. For each image, perturbations are shown on $\mathbf{x}_0$. a) shows the $2$-cp set ($\darkredbox$). b) shows the intersection of $2$-cp set with the adversarial perturbation set for $\mathbf{x}_0$ ($\redbox$). c) shows the adversarial perturbation set for all inputs in the batch ($\square$, $\redbox$). d) shows the adversarial perturbation set for all inputs except for $\mathbf{x}_0$ ($\square$)}
    \label{fig:bounding}
\end{figure*} 

$\mathcal{L}_{\text{$2$CP}}$ is still not usable for training as it is expensive to compute $\mathcal{C}(2,\epsilon)$ (which would require intersecting single-input adversarial regions \cite{dimitrov2021provably}). To make this usable, we first show that for a given input $\mathcal{L}_{\text{$2$CP}}$ can be upper bounded by computing the maximum loss over the $1$-cp perturbation sets from the other inputs in the batch. Formally, we have

\begin{theorem}\label{th:c2} Given $\mathcal{X}_B\subseteq\mathbb{R}^{\din}\times\mathbb{Z}$, a network $f:\mathbb{R}^{\din}\to\mathbb{R}^{\dout}$, a given input $(\mathbf{x}_0, y_0)$, and norm-bound $\epsilon \in \mathbb{R}$. Then,
\begin{equation*}
\mathcal{L}_{2\text{CP}}(\mathcal{X}_B, \mathbf{x}_0, y_0, \epsilon) \leq \max_{\mathcal{C}_{\mathbf{u} \in \mathcal{X}/(\mathbf{x}_0,y_0),f}(1, \epsilon)} \mathcal{L}(f(\mathbf{x}_0 + \mathbf{u}), y_0)
\end{equation*}
\end{theorem}

\begin{sproof}
\cref{fig:bounding} gives a simplified visual guide for our proof, where $|\mathcal{X}_B| =5$. The light red region ($\redbox$) represents the set of adversarial perturbations for $\mathbf{x}_0$. In \cref{fig:bounding} a) we see the $2$-cp set ($\darkredbox$) on top of $\mathbf{x}_0$, $\mathcal{L}_{\text{$2$CP}}$ maximizes the loss over $\darkredbox$. Assuming that we have a standard loss function where adversarial perturbations ($\redbox$) have greater loss than safe perturbations ($\greenbox$) ($\forall \mathbf{v}, \mathbf{v}'\in \mathcal{B}(\mathbf{0}, \epsilon). \neg A_f(\mathbf{x}_0 + \mathbf{v}, y) \wedge  A_f(\mathbf{x}_0 + \mathbf{v}', y) \implies \mathcal{L}(f(\mathbf{x}_0 + \mathbf{v}), y_0) \leq \mathcal{L}(f(\mathbf{x}_0 + \mathbf{v}'), y_0)$), then for $\mathbf{x}_0$ the maximum loss over $\darkredbox$ (LHS) occurs where $\darkredbox$ and $\redbox$ intersect (\cref{fig:bounding} b). We can compute an upper bound on the LHS by maximizing over an overapproximation of the intersection set. This overapproximation can be computed by considering all the single-input adversarial perturbation sets (\cref{fig:bounding} c). Finally, a key observation is that since we are trying to overapproximate the intersection set we do not have to include the set of adversarial perturbations for $\mathbf{x}_0$ reducing overregularization (\cref{fig:bounding} d). This loss in overregularization is significant as seen in \cref{sec:siboxes}. The formal proof can be found in Appendix 
A.2.    
\end{sproof}

Although we cannot exactly compute $\mathcal{C}_{\mathcal{X}/(\mathbf{x},y),f}(1, \epsilon)$, SABR \cite{sabr} shows that we can approximate this set in certified training with small boxes around a precomputed adversarial perturbation. This observation leads to our final loss. 

\begin{definition}
For batch $\mathcal{X}_B$, current input $(\mathbf{x}_i, y_i)\in \mathcal{X}_B$, norm-bound $\epsilon \in \mathbb{R}$, and small box norm-bound $\tau \in \mathbb{R}$, we define the \method loss as 
    \begin{equation*}
    \mathcal{L}_{\text{\method}}(\mathcal{X}_B, \mathbf{x}_i, y_i,  \epsilon, \tau) \coloneqq \sum_{(\mathbf{x}_j,y_j)\in \mathcal{X}_B, i \neq j} \mathcal{L}_{\text{IBP}}(\mathbf{x}_i + \argmax_{\mathbf{v} \in \mathcal{B}(\mathbf{0}, \epsilon - \tau)} \mathcal{L}(\mathbf{x}_j,y_j) , y_i, \tau) 
\end{equation*}
\end{definition}

\begin{figure*}[t]
    \centering
\includegraphics[width=1\textwidth]{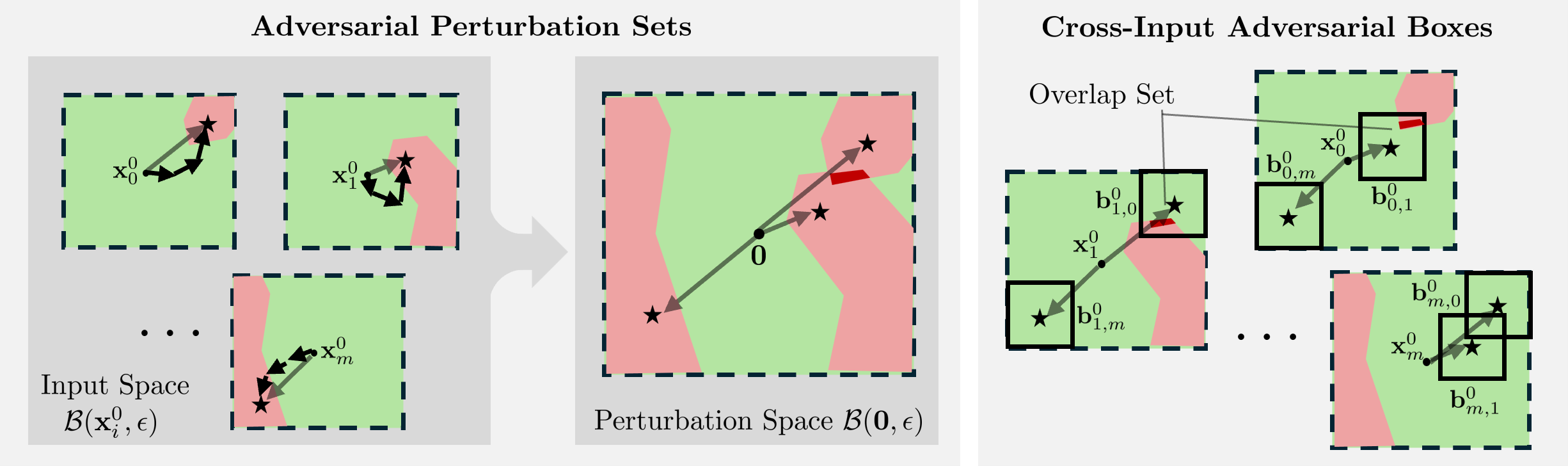}
    \caption{Intuition behind \method. UAPs occur where adversarial regions ($\redbox$) from multiple inputs overlap ($\darkredbox$). To approximate this, adversarial examples ($\star$) are computed for each input, $\mathbf{x}_i$, the corresponding perturbation vectors, $\mathbf{v}_i$, (dark gray \protect\grayarrow{2pt}) and adversarial regions are collocated to $\mathcal{B}(\mathbf{0}, \epsilon)$. To approximate $\darkredbox$ we take cross-input adversarial perturbations and draw $l_\infty$ balls around them, $b_{i,j}^0 = \mathcal{B}(\mathbf{x}_i + \mathbf{v}_j, \tau)$.}
    \label{fig:intuition}
\end{figure*}

\subsection{\method Algorithm}

\noindent\textbf{From Loss to Algorithm}. \Cref{fig:intuition} provides a simplified example to aid visual intuition for the \method algorithm. On the LHS, we show the exact adversarial perturbation sets ($\redbox$) for each input in the batch and view these sets collocated on the origin. As discussed in \cref{sec:citrusloss}, it is expensive to compute the exact adversarial sets \cite{dimitrov2021provably}. Instead, on the RHS, we approximate these regions with small bounding boxes ($\square$) around cross-input adversarial perturbations (not including the same-input perturbation).

\begin{figure*}[t]
    \centering
\includegraphics[width=1\textwidth]{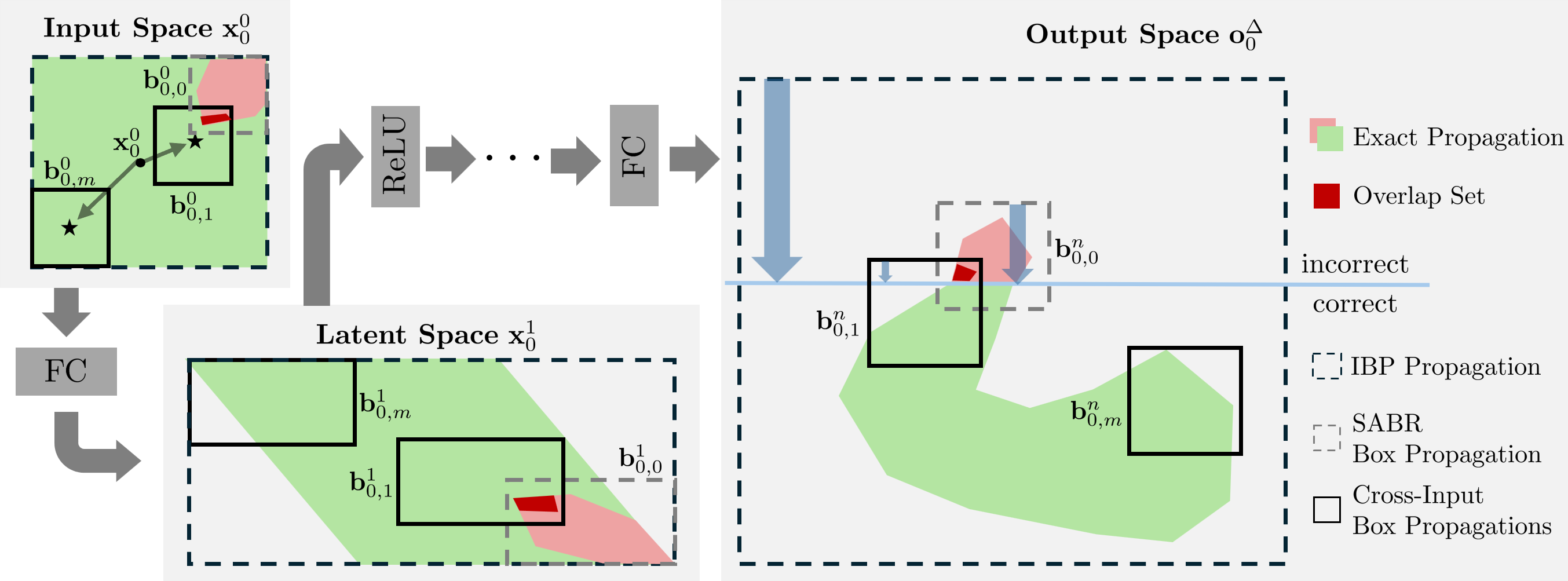}
    \caption{Propagation for $\mathbf{x}_0$ and bounding boxes $b_{0,1}^0$ and $b_{0,m}^0$. The boxes are propagated using IBP through each layer of the network. In the output layer, we see that although $b_{0,1}^n$ does not capture the entire adversarial region for $\mathbf{x}_0$ it does capture the overlapping adversarial region between $\mathbf{x}_0$ and $\mathbf{x}_1$. Compared to IBP propagation ($\dashedbox$) we induce much less regularization (big blue \protect\bigbluearrow{4pt} vs. small blue \protect\smallbluearrow{1pt}). Depending on the size of the overlap region, SABR ($\color{gray}{\dashedbox}$) incurs more regularization (medium blue \protect\bluearrow{2pt}).}
    \label{fig:propagation}
\end{figure*}

\noindent\textbf{Box Propagation}. We illustrate the propagation process for one input, $\mathbf{x}_0$, in \cref{fig:propagation}. All boxes can be propagated through the network using any symbolic propagation method \cite{deeppoly, acrown, gowal2018effectiveness}, but similarly to SABR we use \textsc{Box} propagation \cite{mirman2018differentiable, gowal2018effectiveness} for its speed and well-behaved optimization problem \cite{jovanovic2022paradox}. In the output space, we visualize the propagation of the cross-input adversarial boxes ($\square$) and the single-input adversarial, or SABR, box ($\color{gray}{\dashedbox}$). The dark red region ($\darkredbox$) represents the common perturbation set between $\mathbf{x}_0$ and $\mathbf{x}_1$. We can see that at the end of the propagation $b_{0,1}^n$ includes the entire overlap set even though it does not include the entire adversarial set for $\mathbf{x}_0$. Even though \method propagates more boxes than SABR, many of these boxes incur no regularization ($b_{0,m}^n$) while other boxes include a smaller portion of the adversarial region ($b_{0,1}^n$) incurring less regularization (medium blue \protect\bluearrow{2pt} vs. small blue \protect\smallbluearrow{1pt}) compared to SABR propagation ($\color{gray}{\dashedbox}$). We also show IBP propagation of the entire $l_\infty$ ball and we see that \method incurs significantly less regularization (big blue \protect\bigbluearrow{4pt}).

\noindent\textbf{\method Algorithm}. In \cref{alg:algorithm}, we show training with \method. For each batch, $\mathcal{X}_B$, we first compute a set of adversarial perturbations $\mathbf{v}_i$ for each input $\mathbf{x}_i \in \mathcal{X}_B$ (Line 4). We then iterate through each input in the batch, $\mathbf{x}_i$ and through each cross-input adversarial perturbation, $\mathbf{v}_j$, not from the current input ($i\neq j$). For each input perturbation pair, we update the loss by computing the IBP loss on a box centered at $\mathbf{x}_i + \mathbf{v}_j$ with radius $\tau$ (Line 8). \method is the first algorithm specialized for certified training against universal perturbations.

\vspace{-10pt}
\begin{algorithm}[h]
   \caption{\method Algorithm}
   \label{alg:algorithm}
\begin{algorithmic}[1]
   \STATE Initialize $\theta$
   \FOR{each Epoch}
   \FOR{$\mathcal{X}_B \subset \mathcal{X}$}
        \STATE Compute  an adversarial perturbation $\mathbf{v}_i$ for $\mathbf{x}_i \in \mathcal{X}_B$
        \STATE $\mathcal{L}_{\text{\method}} \gets 0$
        \FOR{$i \in [|\mathcal{X}_B|]$}
            \FOR{$j \in [|\mathcal{X}_B|], j \neq i$}
                \STATE $\mathcal{L}_{\text{\method}} = \mathcal{L}_{\text{\method}} + \mathcal{L}_{\text{IBP}}(\mathbf{x}_i + \mathbf{v}_j, y_i, \tau)$
            \ENDFOR
        \ENDFOR
        \STATE Update $\theta$ using $\mathcal{L}_{\text{\method}}$
   \ENDFOR
   \ENDFOR
\end{algorithmic}
\end{algorithm}
\vspace{-15pt}

\subsection{Same-Input vs. Cross-Input Adversarial Boxes}\label{sec:siboxes}

\cref{fig:bounding} and \ref{fig:propagation} give us some insight into why we do not propagate the same-input adversarial box. To measure the impact of adding same-input adversarial boxes to the loss, we record the average loss across the last epoch of training for a convolutional network trained on CIFAR-10 with $\epsilon = 8/255$ and a batch size of $5$. We compare the loss from the same-input adversarial box (SI) with the sum ($\sum (\text{CI})$) and max loss (max(CI)) from the 4 cross-input adversarial boxes. 
\begin{wrapfigure}{r}{0.55\textwidth}
  \vspace{-20pt}
  \begin{center}
    \includegraphics[width=0.45\textwidth]{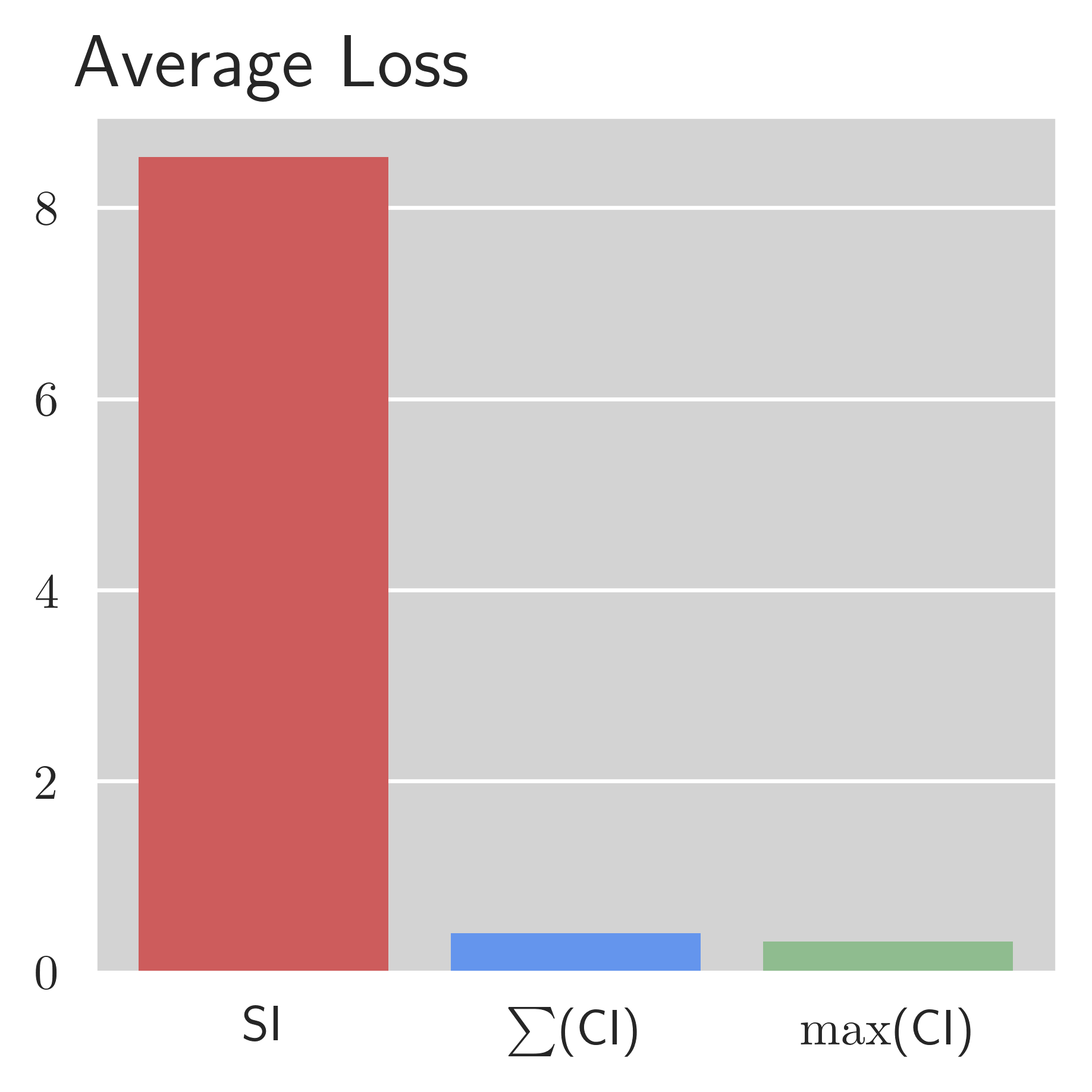}
  \end{center}
  \vspace{-20pt}
  \caption{Comparison of average loss across the last epoch of training incurred by same-input adversarial boxes to the sum and max of cross-input adversarial boxes. Network trained on CIFAR-10 with $\epsilon = \frac{8}{255}$ and batch size of 8.}
  \vspace{-20pt}
\label{fig:ablation_loss}
\end{wrapfigure}
In \cref{fig:ablation_loss} we see that the loss from the same-input adversarial box dominates the loss from the other boxes substantially. This indicates that adding same-input boxes likely incurs significant regularization and reduction in standard accuracy. This is corroborated by our ablation study in \cref{sec:ablation} where we find that adding the same-input adversarial boxes to the training process reduces the model's standard accuracy. We further observe that for the cross-input adversarial boxes, the max loss is close to the sum indicating that a single large loss typically dominates the sum.

%% file: contents/4-evaluation.tex
\section{Evaluation}\label{sec:eval}

We implemented \method in Python \cite{python} and PyTorch \cite{paszke2019pytorch}. 
We compare \method to existing SOTA single-input certified training methods and existing robust UAP training techniques. We also perform ablation studies.

\noindent\textbf{Experimental Setup}. We consider three popular image recognition datasets for certified training: MNIST \cite{mnist}, CIFAR-10 \cite{cifar}, and TinyImageNet \cite{tinyimagenet}. We use a variety of challenging $l_\infty$ perturbation bounds common in verification/robust training literature \cite{acrown, bcrown, deeppoly, deepz, shiibp, sabr, taps}. Unless otherwise indicated, we use a 7-layer convolutional architecture, CNN7, used in many prior works we compare against \cite{shiibp, sabr, taps}. All experiments were performed on a desktop PC with a GeForce RTX(TM) 3090 GPU and a 16-core Intel(R) Core(TM) i9-9900KS CPU @ 4.00GHz. We use a modification on Zeng \etal \cite{iclruapcert} for worst-case UAP certification (note that this is an incomplete verifier so all results are underapproximations of true certified accuracy). Further evaluation, training/verification times, and training parameters can be found in Appendix 
C.

\noindent\textbf{Certified Average UAP Accuracy}. Zeng \etal \cite{iclruapcert} introduce a verifier for UAP certification that can compute a lower bound on the worst-case accuracy under a UAP attack. RaVeN \cite{raven} and RACoon \cite{racoon} further improve upon the verification precision for worst-case UAP accuracy, for this paper we evaluate our results using RACoon. We compute this bound on the worst-case accuracy for sets of 5 images at a time and report the average value across the entire test dataset: we refer to this metric as certified average UAP accuracy. Further details on this metric can be found in Appendix 
B.

\subsection{Main Results}
We compare \method to SOTA certified training methods in \cref{table:main}. Existing certified training methods train for single-input adversarial robustness. For each method, we report the best results achieved under any architecture presented in the respective paper. 

Across all datasets and $\epsilon$s we observe that \method obtains better standard accuracy than existing methods (up to 10.3\% increase for CIFAR-10 $\epsilon = 8/255$). \method obtains SOTA performance for certified average UAP accuracy, obtaining better performance in 3 out of 5 cases than existing baselines (i.e. \method obtains 26.27\% certified average UAP accuracy on TinyImageNet vs. 24.71\% from TAPS \cite{taps}) and comparable performance (\method's certified average UAP accuracy is within 1.69\% of SOTA baselines) on the rest. Our results show that \method indeed decreases regularization and increases standard accuracy while maintaining good certified average UAP accuracy.

\addtolength{\tabcolsep}{1.5pt}
\begin{table*}[t]
\caption{Comparison of standard accuracy (Std) and certified average UAP accuracy (UCert) for different certified training methods on the full MNIST, CIFAR-10, and TinyImageNet test sets. A variation on \cite{iclruapcert} is used for certified average UAP accuracy.}
\vspace{-5pt}
\begin{center}
\begin{footnotesize}
\begin{tabular}{lllccc}
\toprule
Dataset & $\epsilon$ & Training Method & Source & Std [\%] & UCert [\%] \\
\midrule
\multirow{8}{*}{MNIST} & \multirow{4}{*}{$0.1$} & IBP & Shi \etal \cite{shiibp} & 98.84& 98.12\\
& & SABR & M{\"u}ller \etal \cite{sabr} & 99.23& 98.37\\
& & TAPS & Mao \etal \cite{taps} & 99.19& \textbf{98.65}\\
& & \method & this work & \textbf{99.27}& 98.41\\
\cmidrule{2-6}
& \multirow{4}{*}{$0.3$} & IBP & Shi \etal \cite{shiibp} & 97.67& 94.76\\
& & SABR & M{\"u}ller \etal \cite{sabr} & 98.75& 95.37\\
& & TAPS & Mao \etal \cite{taps} & 98.53& 95.24\\
& & \method & this work & \textbf{99.04}& \textbf{95.61}\\
\midrule
\multirow{8}{*}{CIFAR-10} & \multirow{4}{*}{$\frac{2}{255}$} & IBP & Shi \etal \cite{shiibp} & 66.84& 59.41\\
& & SABR & M{\"u}ller \etal \cite{sabr} & 79.24& 65.38\\
& & TAPS & Mao \etal \cite{taps} & 79.76& 66.62\\
& & \method & this work & \textbf{83.45} & \textbf{66.98}\\
\cmidrule{2-6}
& \multirow{4}{*}{$\frac{8}{255}$} & IBP & Shi \etal \cite{shiibp} & 48.94& 39.05\\
& & SABR & M{\"u}ller \etal \cite{sabr} & 52.38& \textbf{41.57}\\
& & TAPS & Mao \etal \cite{taps} & 52.82& 40.90\\
& & \method & this work & \textbf{63.12}& 39.88\\
\midrule
\multirow{4}{*}{TinyImageNet} & \multirow{4}{*}{$\frac{1}{255}$} & IBP & Shi \etal \cite{shiibp} & 25.92& 18.50\\
& & SABR & M{\"u}ller \etal \cite{sabr} & 28.85& 21.53\\
& & TAPS & Mao \etal \cite{taps} & 28.98& 24.71\\
& & \method & this work & \textbf{35.62}& \textbf{26.27}\\
\bottomrule
\end{tabular}
\label{table:main}
\end{footnotesize}
\end{center}
\vspace{-20pt}
\end{table*}
\addtolength{\tabcolsep}{-1.5pt}

\subsection{Comparison to Universal Adversarial Training}

In \cref{table:at} we compare the performance of \method to universal and standard adversarial training methods. We compare to standard single-input PGD adversarial training \cite{pgd}. Shafahi \etal \cite{uat} introduce Universal Adversarial Training (UAT), an efficient way to perform adversarial training against UAPs. Benz \etal \cite{cwuat} introduces a class-wise variant of the UAT algorithm which improves performance. We compare against all of these methods in \cref{table:at} on CIFAR-10 with $\epsilon = 8/255$. We observe that while these methods obtain good standard accuracy (94.91\% for CW-UAT compared to 63.12\% for \method) they perform poorly for certified average UAP accuracy (1.51\% for CW-UAT compared to 39.88\% for \method). Our observation for the certified UAP accuracy of adversarial training is consistent with other studies which show that the standard certified accuracy of adversarial training based methods is low \cite{mirman2018differentiable}.

\addtolength{\tabcolsep}{1.5pt}
\begin{table*}[t]
\caption{Comparison of standard accuracy (Std) and certified average UAP accuracy (UCert) for different universal and standard adversarial training methods on the full CIFAR-10 test sets. A variation on \cite{iclruapcert} is used for certified worst-case UAP accuracy.}
\begin{center}
\begin{footnotesize}
\begin{tabular}{lllccc}
\toprule
Dataset & $\epsilon$ & Training Method & Source & Std [\%] & UCert [\%] \\
\midrule
\multirow{4}{*}{CIFAR-10} & \multirow{4}{*}{$\frac{8}{255}$} & PGD & Madry \etal \cite{pgd} & 87.25& 0.0\\
& & UAT & Shafahi \etal \cite{uat} & 94.28 & 0.87\\
& & CW-UAT & Benz \etal \cite{cwuat} & \textbf{94.91}& 1.51\\
& & \method & this work & 63.12& \textbf{39.88}\\
\bottomrule
\end{tabular}
\label{table:at}
\end{footnotesize}
\end{center}
\end{table*}
\addtolength{\tabcolsep}{-1.5pt}

\begin{figure}
\vspace{-5pt}
\centering
\begin{subfigure}{.48\textwidth}
  \centering
  \includegraphics[width=\textwidth]{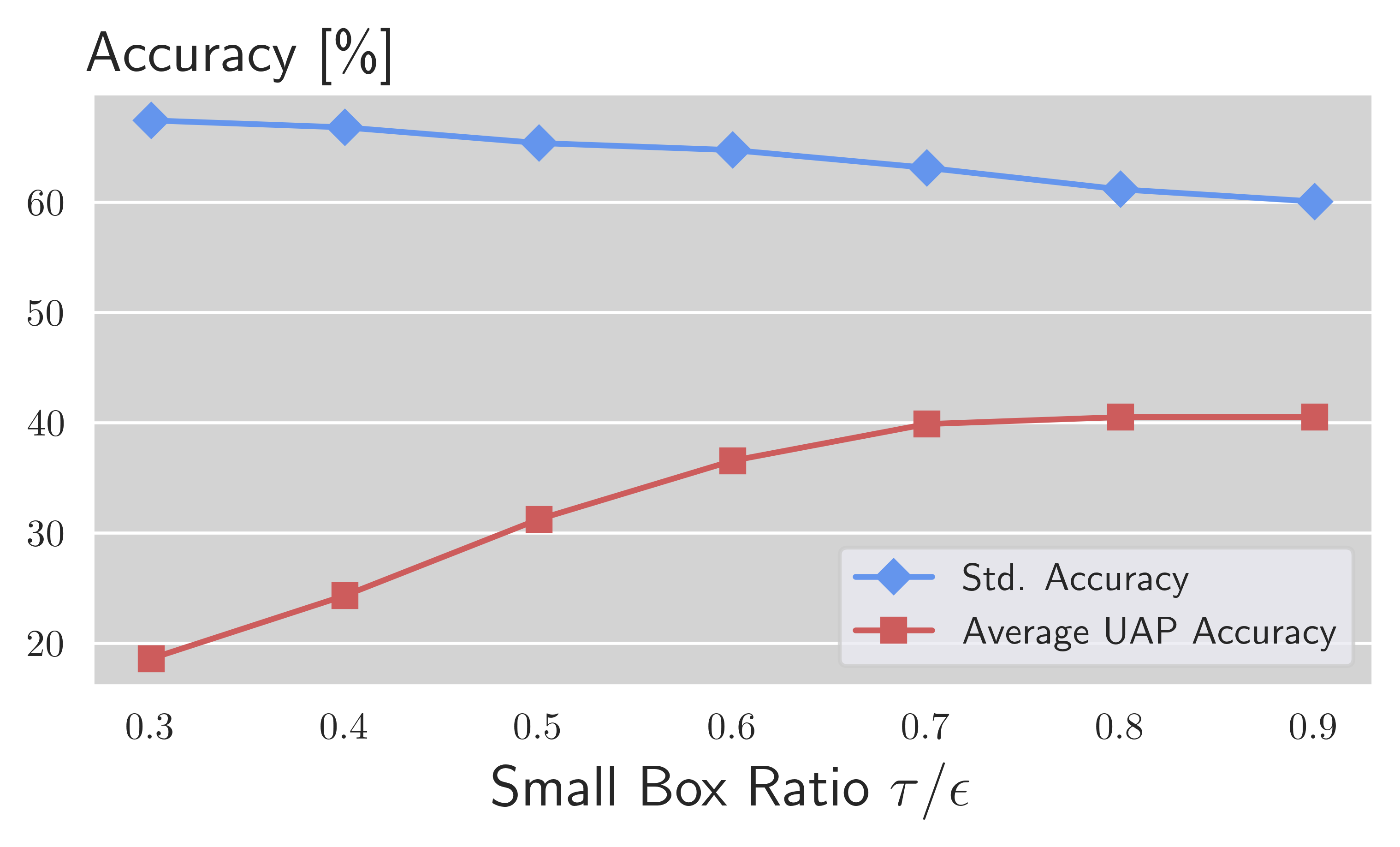}
  \caption{Comparison of standard accuracy and certified average UAP accuracy across different ratios of $\tau/\epsilon$ for \method on CIFAR-10 with $\epsilon = \frac{8}{255}$.}
  \label{fig:ablation_tau}
\end{subfigure}%
\hfill
\begin{subfigure}{.48\textwidth}
  \centering
  \includegraphics[width=\textwidth]{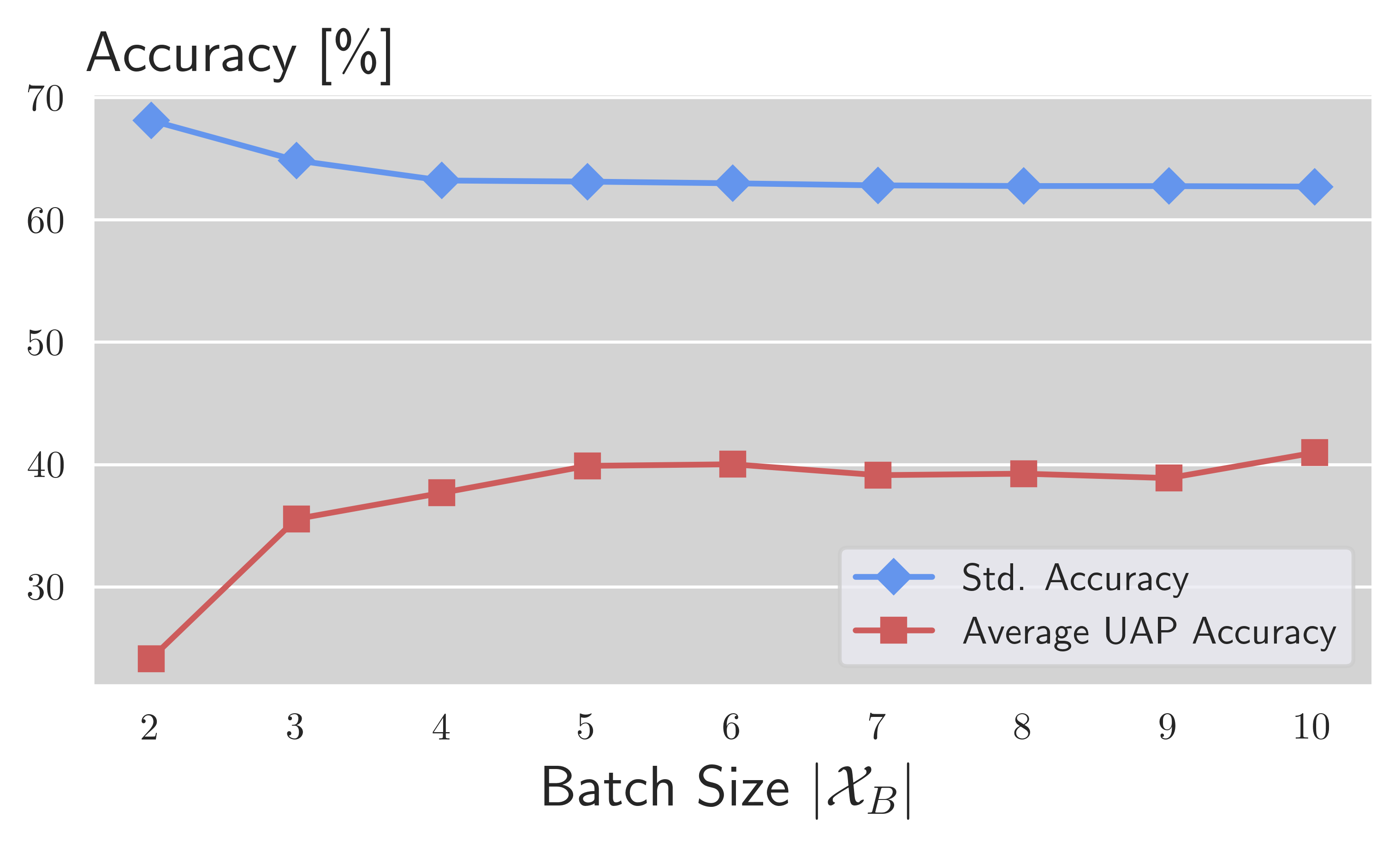}
   \caption{Comparison of standard accuracy and certified average UAP accuracy across different batch sizes for \method on CIFAR-10 with $\epsilon = \frac{8}{255}$.}
  \label{fig:ablation_bs}
\end{subfigure}
\caption{Ablation studies on values of $\tau$ and batch size.}
\label{fig:ablation}
\vspace{-10pt}
\end{figure}

\subsection{Ablation Studies}\label{sec:ablation}
\noindent \textbf{Propagation Region Size}. To study the effect of different values of $\tau$ we vary the ratio of $\tau/\epsilon \in [0.3, 0.9]$ keeping all other training parameters constant. We perform our experiment on CIFAR-10 with $\epsilon = \frac{8}{255}$. In \cref{fig:ablation_tau}, we see that increasing the ratio tends to decrease the final standard accuracy and increases the certified average UAP accuracy. Larger boxes capture more of the true adversarial region but they also incur more regularization.

\noindent\textbf{Training Batch Size}. \method trains each input to be robust on the adversarial regions coming from other inputs in the batch. To study the effect of batch size on overall training performance, we vary the batch size from $2$ to $10$ keeping all other training parameters constant. We perform our experiment on CIFAR-10 with $\epsilon = \frac{8}{255}$. In \cref{fig:ablation_bs}, we see that increasing the batch size tends to decrease the standard accuracy but increases the certified average UAP accuracy. A batch size of $2$ means that each input only sees the adversarial region from $1$ other input, so the certified average UAP accuracy is low. This quickly increases then stabilizes. We note that runtime scales quadratically with the size of the batch.

\noindent\textbf{Same-Input Adversarial Boxes}. To compare the effect of adding same-input adversarial boxes to \method training, visualized in \cref{fig:bounding} c) compared to \cref{fig:bounding} d), we train a network on CIFAR-10 with $\epsilon = 8/255$ using \method but include same-input adversarial boxes while training. In \cref{tab:siab}, we observe that adding 
\begin{wraptable}{r}{0.4\textwidth}
\vspace{-25pt}
\caption{Comparing the effect of adding same-input adversarial boxes (\method + SI) to \method on CIFAR-10 with $\epsilon = \frac{8}{255}$.}
 \begin{tabular}{lcc}
    \toprule
    Method & Std [\%] & UCert [\%] \\
    \midrule
    \method & \textbf{63.12}& 39.88\\
    \method + SI& 50.24& \textbf{40.06}\\
    \bottomrule
    \end{tabular}
\label{tab:siab}
\end{wraptable}
same-input adversarial boxes results in a severe reduction in standard accuracy and a slight increase in certified average UAP accuracy. \method + SI has similar performance to SABR. \cref{fig:ablation_loss} shows the average loss coming from same-input adversarial boxes dominates the loss coming from cross-input adversarial boxes which explains why adding same-input adversarial boxes to \method yields similar performance to SABR.

%% file: contents/5-related_work.tex
\section{Related Work}\label{sec:relatedwork}

\textbf{DNN Verification}. Neural network verification is generally NP-complete \cite{katz2017reluplex} so most existing methods trade precision for scalability. Existing work on DNN verification primarily focuses on single-input robustness verification. Single-input robustness can be deterministically analyzed via abstract interpretation \cite{gowal2018effectiveness, deeppoly, deepz} or via optimization using linear programming (LP) \cite{DePalma2021, prima}, mixed integer linear programming (MILP) \cite{refinezono, tjeng2018evaluating}, or semidefinite programming (SDP) \cite{dathathri2020enabling, raghunathan2018semidefinite}. UAP verification has been proposed using optimization with MILP \cite{iclruapcert}. In this work, we build off of the techniques from abstract interpretation and validate our UAP certification results using a variation on Zeng \etal \cite{iclruapcert}.

\noindent\textbf{Certified Training}. In \cref{sec:background} we primarily focus on certified training techniques using IBP \cite{gowal2018effectiveness, mirman2018differentiable} and unsound improvements on IBP training \cite{sabr}. Zhang \etal propose CROWN-IBP which combines fast \textsc{Box} propagation with tight linear relaxation based bounds in a backwards pass to bound the worst-case loss. Shi \etal \cite{shiibp} show faster IBP training times through a specialized weight initialization method and by adding batchnorm to the layers. Balunovi{\'c} \etal \cite{colt} introduce COLT which integrates an adversary which tries to find points that break verification. Mao \etal \cite{taps} introduces TAPS, a method for combining adversarial (PGD) training with certified (IBP) training by propagating bounds partially through the network then performing PGD training on the remainder of the network. Overall, all existing certified training methods focus on single-input adversarial robustness, whereas to the best of our knowledge, \method is the first method to consider certified training for universal perturbation robustness. 

\noindent \textbf{Randomized smoothing (RS).} RS \cite{randomizedsmoothing, salman2019provably} gives probabilistic certificates of robustness by constructing a smoothed model (different from a DNN and with higher inference cost) for single-input adversaries. Salman \etal \cite{salman2020denoised} design a custom-trained denoiser and use RS to turn any pretrained classifier provably robust; further works use autoencoder \cite{zhang2022robustify} and diffusion \cite{carlinicertified} based denoisers to improve results. In contrast, we do not construct a smoothed model and target deterministic DNN robustness against universal perturbations.

\noindent\textbf{UAP Adversarial Training}. Robust optimization has risen as the primary way to get empirical robustness for DNNs. P{\'e}rolat \etal \cite{perolat2018playing} propose viewing robustness against UAPs as a two player game and propose a method similar to adversarial training but for UAPs. Shafahi \etal \cite{uat} propose universal adversarial training (UAT), a faster method for UAP training in which gradient descent iterations alternate between learning network weights and updating a universal perturbation. Benz \etal \cite{cwuat} improves on UAT by making the observation that UAPs do not attack all classes equally and thus they split the UAT training to be classwise. These empirical robustness methods beat \method when it comes to standard accuracy but they do not obtain much certified average UAP accuracy. 

%% file: contents/6-conclusion.tex
\section{Limitations and Ethics}

In \cref{sec:setloss}, we introduce a set of upper bounding losses for the UAP robustness problem. \method uses $2$-cp loss as we found it offers a good balance between tractable loss computation and standard/certified accuracy tradeoff. However, it may be possible to use tighter upper bounds although the effect on training time and standard/certified accuracy is unknown. We also note that we approximate adversarial regions with boxes around precomputed adversarial examples, while this method was effective it may not be the most precise/best approximation.

\method assumes the UAP threat model which is weaker than the single-input adversarial perturbation threat model considered by existing certified training works. While we argue that the UAP threat model is more realistic in practice, depending on the application area, it could give practitioners a false sense of security if single-input attacks are feasible. 

\section{Conclusion}

We introduce the first certified training method called \method for universal perturbations. \method is based on our theoretical analysis of the UAP robustness problem and based on the key insight that focusing on common adversarial perturbations between $2$ inputs leads us to good certified UAP robustness as well as better standard accuracy than single-input certified training methods. \method lays the groundwork for a new set of certified training methods for universal perturbations.

%% file: contents/7-appendix.tex
\appendix

\input{contents/appendix/proofs}
\input{contents/appendix/eval_details}

%% file: contents/appendix/proofs.tex
\section{Proofs}

In this section, we provide formal proofs for theorems in Sections 
3 and 
4.

\subsection{Proofs for $k$-Common Perturbation Bounding} \label{app:kcpproof}

In this section, we provide a formal proof for Theorem 
1.

\begin{lemma} \label{lm:kcp}
    Given $(\mathbf{x}, y)\in\mathcal{X}$, network $f:\mathbb{R}^{\din}\to\mathbb{R}^{\dout}$, norm-bound $\epsilon \in \mathbb{R}$, $\mathbf{u}^\ast$ as defined in Eq. 
    5, and $k \in \mathbb{N}$ s.t. $\Psi(\mathbf{u}^\ast,k)$, then
    \begin{equation*}
        \mathcal{L}(f(\mathbf{x} + \mathbf{u}^\ast), y) \leq \max_{\mathbf{u}\in\mathcal{C}_{\mathcal{X}, f}(k, \epsilon)}\mathcal{L}(f(\mathbf{x} + \mathbf{u}), y)
    \end{equation*}
\end{lemma}

\begin{proof}
    $\mathbf{u}^\ast \in \mathcal{B}(\mathbf{0}, \epsilon)$ by definition of $\mathbf{u}^\ast$. $\Psi(\mathbf{u}^\ast,k)$ by definition of $k$. Therefore, $\mathbf{u}^\ast \in \mathcal{C}_{\mathcal{X}, f}(k, \epsilon)$. The statement of the lemma then follows by definition of max as $\forall i\in[j]. x_i \leq \max(x_1, \dots, x_j)$.
\end{proof}

\begin{lemma}\label{lm:kjcp}
 Given $(\mathbf{x}, y)\in\mathcal{X}$, network $f:\mathbb{R}^{\din}\to\mathbb{R}^{\dout}$, norm-bound $\epsilon \in \mathbb{R}$, $\mathbf{u}^\ast$ as defined in Eq. 
 5, and $k,j \in \mathbb{N}$ s.t. $k < j$, then
    \begin{equation*}
        \max_{\mathbf{u}\in\mathcal{C}_{\mathcal{X}, f}(j, \epsilon)}\mathcal{L}(f(\mathbf{x} + \mathbf{u}), y) \leq \max_{\mathbf{u}\in\mathcal{C}_{\mathcal{X}, f}(k, \epsilon)}\mathcal{L}(f(\mathbf{x} + \mathbf{u}), y)
    \end{equation*}
\end{lemma}

\begin{proof}
    If $\mathbf{u}\in\mathcal{C}_{\mathcal{X}, f}(j, \epsilon)$ then we have $\mathbf{u} \in \mathcal{B}(\mathbf{0}, \epsilon)$ and $\Psi(\mathbf{u}, j) = 1$. $\forall i < j. \Psi(\mathbf{u}, j) \implies \Psi(\mathbf{u}, i)$ as the existence of a tuple of size $j$ which is all misclassified by $u$ means that all subsets of that tuple (sizes $i < j$) are misclassified by $u$. Therefore, $\forall k < j. \mathbf{u}\in\mathcal{C}_{\mathcal{X}, f}(j, \epsilon)\implies \mathbf{u}\in\mathcal{C}_{\mathcal{X}, f}(k, \epsilon)$ and the statement of the lemma follows through the definition of max.
\end{proof}



\begin{theorem}
    Given $\mathcal{X}\subseteq \mathbb{R}^{\din}\times\mathbb{N}$, network $f:\mathbb{R}^{\din}\to\mathbb{R}^{\dout}$, $\mathbf{u}^\ast$ as defined in Equation 
    5, and norm-bound $\epsilon \in \mathbb{R}$. Let $\kappa^\ast = \hat{\Psi}_{\mathcal{X}, f}(\mathbf{u}^\ast)$ and 
    
    $$\mathcal{E}(k, \epsilon) = \mathop{\mathbb{E}}_{(x,y) \in \mathcal{X}}\left[ \max_{\mathbf{u}\in\mathcal{C}_{\mathcal{X}, f}(k, \epsilon)}\mathcal{L}(f(\mathbf{x} + \mathbf{u}), y)\right]$$
    
    then,
    \begin{equation*}
        \max_{\mathbf{u}\in \mathcal{B}(0,\epsilon)}\left(\mathop{\mathbb{E}}_{(x,y) \in \mathcal{X}}\left[ \mathcal{L}(f(\mathbf{x} + \mathbf{u}), y)\right]\right)\leq \mathcal{E}(\kappa^\ast, \epsilon) \leq \mathcal{E}(\kappa^\ast - 1, \epsilon) \leq \dots \leq \mathcal{E}(1, \epsilon)
    \end{equation*}
\end{theorem}
\begin{proof}
\begin{align*}
    \max_{\mathbf{u}\in \mathcal{B}(0,\epsilon)}\left(\mathop{\mathbb{E}}_{(x,y) \in \mathcal{X}}\left[ \mathcal{L}(f(\mathbf{x} + \mathbf{u}), y)\right]\right)&= \mathop{\mathbb{E}}_{(x,y) \in \mathcal{X}}\left[ \mathcal{L}(f(\mathbf{x} + \mathbf{u}^\ast), y)\right] && (\text{by Eq. 5})\\
     &\leq \mathop{\mathbb{E}}_{(x,y) \in \mathcal{X}}\left[ \max_{\mathbf{u}\in\mathcal{C}_{\mathcal{X}, f}(\kappa^\ast, \epsilon)}\mathcal{L}(f(\mathbf{x} + \mathbf{u}), y)\right] && (\text{by Lm. \ref{lm:kcp}})\\
    &= \mathcal{E}(\kappa^\ast, \epsilon) && (\text{by Def})\\
    &\leq \mathcal{E}(\kappa^\ast - 1, \epsilon) && (\text{by Lm. \ref{lm:kjcp}})\\
    &\dots\\
    &\leq \mathcal{E}(1, \epsilon) && (\text{by Lm. \ref{lm:kjcp}})\\
\end{align*}
\end{proof}

\subsection{Proofs for $\mathcal{L}_{\text{\method}}$ being an upper bound for $\mathcal{L}_{2\text{CP}}$} \label{app:citrusproof}

In this section, we provide a formal proof for Theorem 
2. For these proofs, we assume a standard loss function where additive perturbations which are adversarial incur greater loss than additive perturbations which are safe, i.e. $\forall \mathbf{v}, \mathbf{v}'\in \mathcal{B}(\mathbf{0}, \epsilon). \neg A_f(\mathbf{x}_0 + \mathbf{v}, y) \wedge  A_f(\mathbf{x}_0 + \mathbf{v}', y) \implies \mathcal{L}(f(\mathbf{x}_0 + \mathbf{v}), y_0) \leq \mathcal{L}(f(\mathbf{x}_0 + \mathbf{v}'), y_0)$.

\begin{definition}
    The \textbf{adversarial set}, $\mathcal{S}_{f}(\mathbf{x}_0,y_0, \epsilon) \subseteq \mathcal{B}(\mathbf{0}, \epsilon)$, for a point $(\mathbf{x}_0,y_0)$ is defined as the set of points in $\mathcal{B}(\mathbf{0}, \epsilon)$ which cause $f$ to misclassify when added to $\mathbf{x}_0$. That is, 
    
    $$\mathcal{S}_{f}(\mathbf{x}_0,y_0, \epsilon)\coloneq \{\mathbf{v}|A_f(\mathbf{x}_0 + \mathbf{v}, y_0) \wedge \mathbf{v}\in\mathcal{B}(\mathbf{0}, \epsilon)\}$$

    Further, let $\neg\mathcal{S}_{f}(\mathbf{x}_0,y_0, \epsilon) \subseteq \mathcal{B}(\mathbf{0}, \epsilon)$ indicate the safe set. That is,

    $$\neg\mathcal{S}_{f}(\mathbf{x}_0,y_0, \epsilon)\coloneq \{\mathbf{v}|\neg A_f(\mathbf{x}_0 + \mathbf{v}, y_0) \wedge \mathbf{v}\in\mathcal{B}(\mathbf{0}, \epsilon)\}$$
\end{definition}

Using the definition of an adversarial set, we can now show that the loss for $(\mathbf{x}_0, y_0)$ over $\mathcal{C}_{\mathcal{X}_B,f}(2, \epsilon)$ must occur in the adversarial set for $(\mathbf{x}_0, y_0)$.

\begin{lemma}\label{lm:adv}
Given $\mathcal{X}_B\subseteq\mathbb{R}^{\din}\times\mathbb{Z}$, a network $f:\mathbb{R}^{\din}\to\mathbb{R}^{\dout}$, a given input $(\mathbf{x}_0, y_0)$, and norm-bound $\epsilon \in \mathbb{R}$. If $\mathcal{C}_{\mathcal{X}_B,f}(2, \epsilon)\cap\mathcal{S}_f(\mathbf{x}_0, y_0, \epsilon)\neq \emptyset$ then,
\begin{equation*}
\mathcal{L}_{2\text{CP}}(\mathcal{X}_B, \mathbf{x}_0, y_0, \epsilon) = \max_{\mathbf{u} \in \mathcal{C}_{\mathcal{X}_B,f}(2, \epsilon) \cap \mathcal{S}_f(\mathbf{x}_0, y_0, \epsilon)} \mathcal{L}(f(\mathbf{x}_0 + \mathbf{u}), y_0)
\end{equation*}
\end{lemma}

\begin{proof}
    By definition we have that,
    \begin{equation*}\label{eq:2cpdef}
    \mathcal{L}_{2\text{CP}}(\mathcal{X}_B, \mathbf{x}_0, y_0, \epsilon) = \max_{\mathbf{u}\in\mathcal{C}_{\mathcal{X}_B, f}(2, \epsilon)}\mathcal{L}(f(\mathbf{x} + \mathbf{u}), y)
    \end{equation*}
    Let $\mathbf{u}'$ be the point which maximizes the RHS of the definition above (there may be multiple points which maximize the RHS; however, without loss of generality assume that $\mathbf{u}'$ is unique), that is
    \begin{equation*}
        \max_{\mathbf{u}\in\mathcal{C}_{\mathcal{X}_B, f}(2, \epsilon)}\mathcal{L}(f(\mathbf{x} + \mathbf{u}), y) = \mathcal{L}(f(\mathbf{x} + \mathbf{u}'), y)
    \end{equation*}
    By the assumption we made for standard loss functions, we have that
    \begin{equation*}
        \max_{\mathbf{u}\in\neg\mathcal{S}_f(\mathbf{x}_0, y_0, \epsilon)}\mathcal{L}(f(\mathbf{x} + \mathbf{u}), y) \leq \max_{\mathbf{u}\in\mathcal{S}_f(\mathbf{x}_0, y_0, \epsilon)}\mathcal{L}(f(\mathbf{x} + \mathbf{u}), y) 
    \end{equation*}
    This implies that
    \begin{equation*}
        \max_{\mathbf{u}\in\mathcal{C}_{\mathcal{X}_B, f}(2, \epsilon)\cap\neg\mathcal{S}_f(\mathbf{x}_0, y_0, \epsilon)}\mathcal{L}(f(\mathbf{x} + \mathbf{u}), y) \leq \max_{\mathbf{u}\in\mathcal{C}_{\mathcal{X}_B, f}(2, \epsilon)\cap\mathcal{S}_f(\mathbf{x}_0, y_0, \epsilon)}\mathcal{L}(f(\mathbf{x} + \mathbf{u}), y) 
    \end{equation*}
    Therefore, since $\mathbf{u}'$ maximizes the loss over $C_{\mathcal{X}_B, f}(2, \epsilon)$ the above inequality implies that $\mathbf{u}' \in C_{\mathcal{X}_B, f}(2, \epsilon)\cap\mathcal{S}_f(\mathbf{x}_0, y_0, \epsilon)$. That is,
    \begin{equation*}
        \mathcal{L}(f(\mathbf{x} + \mathbf{u}'), y) = \max_{\mathbf{u}\in\mathcal{C}_{\mathcal{X}_B, f}(2, \epsilon)\cap\mathcal{S}_f(\mathbf{x}_0, y_0, \epsilon)}\mathcal{L}(f(\mathbf{x} + \mathbf{u}), y)
    \end{equation*}
    Which gives the statement of the lemma with the definition of $\mathbf{u}'$.
\end{proof}

\begin{definition}\label{def:intersect}
    We define the \textbf{cross-input adversarial set}, $\mathcal{CI}_f(\mathcal{X}_B, \mathbf{x}_0, y_0, \epsilon)$, to be the intersection between the adversarial sets for all points in $\mathcal{X}_B$ besides $(\mathbf{x}_0, y_0)$. That is,
    
    $$\mathcal{CI}_f(\mathcal{X}_B, \mathbf{x}_0, y_0, \epsilon) \coloneq \bigcup_{(\mathbf{x}_i, y_i) \in \mathcal{X}_B, \mathbf{x}_i \neq \mathbf{x}_0} \mathcal{S}_f(\mathbf{x}_i, y_i, \epsilon)$$

    Note that, $\mathcal{CI}_f(\mathcal{X}_B, \mathbf{x}_0, y_0, \epsilon) = \mathcal{C}_{\mathcal{X}_B/(\mathbf{x}_0,y_0),f}(1, \epsilon)$.
\end{definition}

Using the definition of $\mathcal{CI}_f(\mathcal{X}_B, \mathbf{x}_0, y_0, \epsilon)$ we can bound the maximum loss for the current input, $(\mathbf{x}_0, y_0)$, over the intersection of $\mathcal{C}_{\mathcal{X}_B,f}(2, \epsilon)$ and $\mathcal{S}_f(\mathbf{x}_0, y_0, \epsilon)$ by maximizing the loss over the cross-input adversarial set for $(\mathbf{x}_0, y_0)$.

\begin{lemma}\label{lm:intersect}
Given $\mathcal{X}_B\subseteq\mathbb{R}^{\din}\times\mathbb{Z}$, a network $f:\mathbb{R}^{\din}\to\mathbb{R}^{\dout}$, a given input $(\mathbf{x}_0, y_0)$, and norm-bound $\epsilon \in \mathbb{R}$. Then,
\begin{equation*}
\max_{\mathbf{u} \in \mathcal{C}_{\mathcal{X}_B,f}(2, \epsilon) \cap \mathcal{S}_f(\mathbf{x}_0, y_0, \epsilon)} \mathcal{L}(f(\mathbf{x}_0 + \mathbf{u}), y_0) \leq \max_{\mathbf{u} \in \mathcal{CI}_f(\mathcal{X}_B, \mathbf{x}_0, y_0, \epsilon)} \mathcal{L}(f(\mathbf{x}_0 + \mathbf{u}), y_0)
\end{equation*}
\end{lemma}

\begin{proof}
    Given $\mathbf{u} \in \mathcal{C}_{\mathcal{X}_B,f}(2, \epsilon) \cap \mathcal{S}_f(\mathbf{x}_0, y_0, \epsilon)$ by definition of $\mathcal{C}_{\mathcal{X}_B,f}(2, \epsilon)$ we know that $\exists \{(\mathbf{x}_i, y_i),(\mathbf{x}_j, y_j)\}\subseteq\mathcal{X}_B.A_f(\mathbf{x}_i + \mathbf{u}, y_i) \wedge A_f(\mathbf{x}_j + \mathbf{u} + y_j)$. By definition of $\mathcal{S}_f(\mathbf{x}_0, y_0, \epsilon)$ we know that one of $\mathbf{x}_i, \mathbf{x}_j$ equals $\mathbf{x}_0$. Thus, we know that there is at least one more input which $\mathbf{u}$ is adversarial for, that is $\exists (\mathbf{x}_i, y_i)\subseteq\mathcal{X}_B, \mathbf{x}_i \neq \mathbf{x_0}.A_f(\mathbf{x}_i + \mathbf{u}, y_i) \wedge A_f(\mathbf{x}_j + \mathbf{u} + y_j )$. This implies that $\exists (\mathbf{x}_i, y_i)\subseteq\mathcal{X}_B, \mathbf{x}_i \neq \mathbf{x_0}.\mathbf{u} \in \mathcal{S}_f(\mathbf{x}_i, y_i, \epsilon)$. If we take the union of all the adversarial sets of all other inputs, this union must also contain $\mathbf{u}$, in other words, $\mathbf{u} \in \mathcal{CI}_f(\mathcal{X}_B, \mathbf{x}_0, y_0, \epsilon)$. This gives us that,

    \begin{equation*}
        \mathbf{u} \in \mathcal{C}_{\mathcal{X}_B,f}(2, \epsilon) \cap \mathcal{S}_f(\mathbf{x}_0, y_0, \epsilon)\implies\mathbf{u}\in \mathcal{CI}_f(\mathcal{X}_B, \mathbf{x}_0, y_0, \epsilon)
    \end{equation*}
    The statement of the lemma follows by the definition of max.
\end{proof}

Using Lemmas \ref{lm:adv} and \ref{lm:intersect} we now have our main result which leads to \method loss.

\begin{theorem} \label{th:appth2} Given $\mathcal{X}_B\subseteq\mathbb{R}^{\din}\times\mathbb{Z}$, a network $f:\mathbb{R}^{\din}\to\mathbb{R}^{\dout}$, a given input $(\mathbf{x}_0, y_0)$, and norm-bound $\epsilon \in \mathbb{R}$. Then, 
\begin{equation*}
\mathcal{L}_{2\text{CP}}(\mathcal{X}_B, \mathbf{x}_0, y_0, \epsilon) \leq \max_{\mathbf{u} \in \mathcal{C}_{\mathcal{X}/(\mathbf{x}_0,y_0),f}(1, \epsilon)} \mathcal{L}(f(\mathbf{x}_0 + \mathbf{u}), y_0)
\end{equation*}
\end{theorem}

\begin{proof}
\begin{case}
    $\mathcal{C}_{\mathcal{X}_B,f}(2, \epsilon)\cap\mathcal{S}_f(\mathbf{x}_0, y_0, \epsilon)\neq \emptyset$
\end{case}
    \begin{align*}
        \mathcal{L}_{2\text{CP}}(\mathcal{X}_B, \mathbf{x}_0, y_0, \epsilon) &= \max_{\mathbf{u} \in \mathcal{C}_{\mathcal{X}_B,f}(2, \epsilon) \cap \mathcal{S}_f(\mathbf{x}_0, y_0, \epsilon)} \mathcal{L}(f(\mathbf{x}_0 + \mathbf{u}), y_0) && (\text{by \cref{lm:adv}})\\
        &\leq \max_{\mathbf{u} \in \mathcal{CI}_f(\mathcal{X}_B, \mathbf{x}_0, y_0, \epsilon)} \mathcal{L}(f(\mathbf{x}_0 + \mathbf{u}), y_0) && (\text{by \cref{lm:intersect}})\\
        &= \max_{\mathbf{u} \in \mathcal{C}_{\mathcal{X}/(\mathbf{x}_0,y_0),f}(1, \epsilon)} \mathcal{L}(f(\mathbf{x}_0 + \mathbf{u}), y_0) && (\text{by \cref{def:intersect}})
    \end{align*}
\begin{case}
    $\mathcal{C}_{\mathcal{X}_B,f}(2, \epsilon)\cap\mathcal{S}_f(\mathbf{x}_0, y_0, \epsilon)= \emptyset$
\end{case}
$\mathcal{C}_{\mathcal{X}_B,f}(2, \epsilon)\cap\mathcal{S}_f(\mathbf{x}_0, y_0, \epsilon)= \emptyset$ implies $(\mathbf{x}_0, y_0)$ is not susceptible to any universal perturbation which affects it and another input at the same time, that is $\forall \mathbf{u} \in \mathcal{B}(\mathbf{0}, \epsilon), (\mathbf{x}_i, y_i)\in\mathcal{X}_B, \mathbf{x}_i \neq \mathbf{x_0}.\neg(A_f(\mathbf{x}_0 + \mathbf{u}, y_0) \wedge A_f(\mathbf{x}_i + \mathbf{u}, y_i))$. In other words, $(\mathbf{x}_0, y_0)$ is already safe from UAPs and training on adversarial sets for other inputs will not incur regularization.
\end{proof}

\section{Evaluation Metric} \label{app:metric}

Ideally, we want to verify the worst-case accuracy due to a single adversarial perturbation across an entire dataset or data distribution; however, this is infeasible. Instead, we verify using RACoon 
[3] (which uses the same metric as RaVeN 
[4] and Zeng \etal [52]). We compute a bound over a set of $N$ images and measure the worst-case performance of a single perturbation on these inputs, i.e. a universal perturbation for that set of images. Note that the average worst-case UAP accuracy for batch size $N$ also lower bounds the true worst-case UAP accuracy of the entire dataset. We formally state and prove this fact below.

\begin{definition}
    $\mathcal{Z}_{f}: \mathbb{R}^{\din} \times \mathbb{N} \to \mathbb{N}$ is a function which takes a dataset, $\mathcal{X}$ and returns the largest $k$ s.t. a k-cp exists on $\mathcal{X}$ and network $f$. That is, $\mathcal{C}_{\mathcal{X}, f}(\mathcal{Z}_{f}(\mathcal{X}), \epsilon) \neq \emptyset$ and $\mathcal{C}_{\mathcal{X}, f}(\mathcal{Z}_{f}(\mathcal{X}) + 1, \epsilon) = \emptyset$.
\end{definition}

\begin{theorem} \label{thm:batch}
     Given a network $f$, a dataset, $\mathcal{X}$ with size $M$ if we split $\mathcal{X}$ into batches of size $N$ s.t. $N|M$, $\mathcal{X}_1, \dots, \mathcal{X}_{M/N}$, then 
    $$1 - \frac{1}{M}\sum_{i=1}^{M/N} \mathcal{Z}_{f}(\mathcal{X}_i) \leq  1 - \frac{\mathcal{Z}_{f}(\mathcal{X})}{M}$$
\end{theorem}

\begin{proof}
    Let $\mathbf{u}^*$ be a perturbation which is a $\mathcal{Z}_{f}(\mathcal{X})$-cp on $\mathcal{X}$ and $f$. Note that by definition of $\mathcal{Z}$ $\mathbf{u}^*$ must exist and there is no $(\mathcal{Z}_{f}(\mathcal{X}) + 1)$-cp on $\mathcal{X}$ and $f$. Since $\mathcal{X}$ is made up from batches $\mathcal{X}_1, \dots, \mathcal{X}_{M/N}$ we have that 

    $$\sum_{i=1}^{M/N}\hat{\Psi}_{\mathcal{X}_i, f}(\mathbf{u}^*) = \mathcal{Z}_f(\mathcal{X})$$
    
    Furthermore, by definition of $\mathcal{Z}$ $\mathbf{u}^*$ is at most a $\mathcal{Z}_{f}(\mathcal{X}_i)$-cp on $\mathcal{X}_i$ and $f$. Thus, we have

    $$\hat{\Psi}_{\mathcal{X}_i, f}(\mathbf{u}^*) \leq \mathcal{Z}_f(\mathcal{X}_i)$$

    Combining these two facts we can write

    $$\mathcal{Z}_f(\mathcal{X}) = \sum_{i=1}^{M/N}\hat{\Psi}_{\mathcal{X}_i, f}(\mathbf{u}^*) \leq \sum_{i=1}^{M/N} \mathcal{Z}_f(\mathcal{X}_i)$$

    Which can be trivially rearranged to the statement of the theorem.
\end{proof}

Furthermore, individual verification (i.e. as computed by methods such as CROWN-IBP 
[53]) is a lower bound for worst-case UAP accuracy. For example, if we are attempting to verify the worst-case UAP accuracy for inputs A, B, and C, we can first individually verify each input. If A can already be verified then we can be sure that no UAP exists for A and ignore it in our computation. In other words, the worst-case UAP accuracy in this case is at least $1/3$ and we can now try to verify if a UAP exists for B and C. This can also be seen by using a batch size of 1 in Theorem \ref{thm:batch}. In fact, Zeng et al. [52] perform individual verification first before UAP verification and they compare their certification to sample-wise (individual) verification results finding that they improve over this metric by up to $12\%$. For the results in our paper, we compare all networks on their certified UAP accuracy.

%% file: contents/appendix/eval_details.tex
\section{Further Experimental Setup Details}\label{app:exp}

In this section, we provide details on experimental setup as well as runtimes.

\subsection{Training and Architecture Details}

We use a batch-size of 5 when training for all experiments in the paper. We use a similar setup to prior works in certified training 
[28,31,40]. Including the weight initialization and regularization from Shi \etal 
[40] and the $\tau/\epsilon$ ratio from M{\"u}ller \etal 
[31]. We used a longer PGD search with $20$ steps (vs $8$ for SABR) when selecting the centers for our propagation regions. 

Similar to prior work 
[28,31,40] we use a 7-layer convolutional DNN, CNN7. The first 5 layers are convolutional with filter sizes of $[64, 64, 128, 128, 128]$, kernel size $3$, strides $[1, 1, 2, 1, 1]$, and padding $1$. Then followed by two fully connected layers of size $512$ and one with size of the output dimension.

\subsection{Training/Verification Runtimes}

\begin{wraptable}{r}{0.4\textwidth}
\vspace{-25pt}
\caption{\method training runtimes.}
 \begin{tabular}{lcc}
    \toprule
    Dataset & $\epsilon$ & Time (mins)\\
    \midrule
    \multirow{2}{*}{MNIST} & 0.1 & 527\\
    & 0.3 & 492\\
    \midrule
    \multirow{2}{*}{CIFAR-10} & 2/255 & 1004\\
    & 8/255 & 1185\\
    \midrule
    TinyImageNet & 1/255 & 3583\\
    \bottomrule
    \end{tabular}
\label{tab:runtimes}
\end{wraptable}
All experiments were performed on a desktop PC with a GeForce RTX(TM) 3090 GPU and a 16-core Intel(R) Core(TM) i9-9900KS CPU @ 4.00GHz. The training runtimes for different datasets and $\epsilon$s can be seen in \cref{tab:runtimes}. Training for times for \method are roughly linear to batch-size and SABR runtimes, for example, SABR takes around $238$ minutes to train on the same hardware for CIFAR-10 and $\epsilon = 8/255$ which is $4.21\times$ less than \method. Verification for \method trained networks on the entire test dataset takes around 8h for MNIST, 11h for CIFAR-10, and 18h for TinyImageNet for each network.